\documentclass{article}
\usepackage{ijcai26}

\usepackage{times}
\usepackage{soul}
\usepackage{url}
\usepackage[hidelinks]{hyperref}
\usepackage[utf8]{inputenc}
\usepackage[small]{caption}
\usepackage{graphicx}
\usepackage{amsmath}
\usepackage{amsthm}
\usepackage{amssymb}
\usepackage{booktabs}
\usepackage{algorithm}
\usepackage{algorithmic}
\usepackage{multirow}
\usepackage{array}
\usepackage{makecell}

\usepackage[dvipsnames]{xcolor}

\title{Uni-RS: A Spatially Faithful Unified Understanding and Generation Model for Remote Sensing}

\author{
Weiyu Zhang$^{1}$ 
\and
Yuan Hu$^{1}$\and
Yong Li$^{1,2}$\And
Yu Liu$^{1}$\\
\affiliations
$^1$Institute of Remote Sensing and Geographical Information System, School of Earth and Space Sciences, Peking University, Beijing, China
\\
$^2$Department of Civil and Environmental Engineering, The Hong Kong University of Science and Technology, Hong Kong
\emails
\{weiyu\_zhang, huyuan\}@pku.edu.cn,
yong.li@connect.ust.hk,
liuyu@urban.pku.edu.cn
}

\begin{document}

\maketitle

\begin{abstract}
Unified remote sensing multimodal models exhibit a pronounced spatial reversal curse: although they can accurately recognize and describe object locations in images, they often fail to faithfully execute the same spatial relations during text-to-image generation, where such relations constitute core semantic information in remote sensing.
Motivated by this observation, we propose Uni-RS, the first unified multimodal model tailored for remote sensing, to explicitly address the spatial asymmetry between understanding and generation.
Specifically, we first introduce explicit Spatial-Layout Planning to transform textual instructions into spatial layout plans, decoupling geometric planning from visual synthesis.
We then impose Spatial-Aware Query Supervision to explicitly bias learnable queries toward spatial relations specified in the instruction.
Finally, we develop Image–Caption Spatial Layout Variation to expose the model to systematic geometry-consistent spatial transformations.
Extensive experiments across multiple benchmarks show that our approach substantially improves spatial faithfulness in text-to-image generation, while maintaining strong performance on multimodal understanding tasks like image captioning, visual grounding, and VQA tasks.
\end{abstract}

\section{Introduction}

The evolution of Multimodal Large Language Models (MLLMs) is shifting from perception-centric architectures towards unified frameworks capable of both understanding and generation~\cite{zhang2025unified}.
In the context of Remote Sensing (RS), this shift is particularly transformative. A unified model can serve as a generative simulator~\cite{yang2025dpmamba,liu2025text2earth}, synthesizing samples for data-scarce scenarios (e.g., disaster aftermaths), while facilitating language-driven image editing (e.g., removing cloud artifacts). While unified multimodal understanding and generation have been widely explored in general domains~\cite{team2024chameleon,xie2025showo,pan2025transfer}, to the best of our knowledge, such a unified framework has not yet been studied in the field of remote sensing.

\begin{figure}[t]
    \centering
    \includegraphics[width=0.95\linewidth]{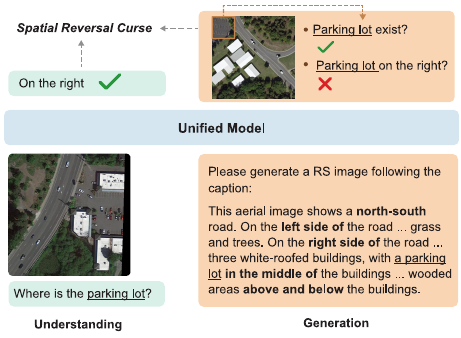}
    \caption{ Illustration of the Spatial Reversal Curse in unified remote sensing multimodal models. While the model correctly describes spatial relations from an input image (left), it often fails to faithfully execute the same relations during text-to-image generation (right).}
    \label{fig:reversal}
\end{figure}

\renewcommand{\thefootnote}{}
\footnotetext{\hspace{-1.8em}The source code is publicly available at \url{https://github.com/zwy-Giser/Uni-RS}. The supplementary materials are available at \url{https://bit.ly/uni_rs_supp}.}
\renewcommand{\thefootnote}{\arabic{footnote}}

However, adapting unified models to the RS domain presents unique challenges. Unlike natural images typically centered on salient objects, RS imagery is characterized by a ``bird's-eye view"~\cite{xia2018dota}, featuring high object density and intricate spatial layouts, where spatial relations are as semantically critical as the objects themselves.
When standard unified models are directly adapted to the remote sensing domain, a critical limitation emerges. Similar to the well-documented \emph{Reversal Curse}~\cite{berglund2024the} in large language models—where knowledge acquired in one direction does not reliably transfer to its inverse—we observe an analogous phenomenon in remote sensing unified models. Specifically, models can accurately recognize and describe spatial layouts from images, yet fail to faithfully reproduce the same layouts during text-to-image generation (Figure \ref{fig:reversal}). We term this systematic asymmetry the \emph{Spatial Reversal Curse}.
For instance, a model that correctly identifies ``a stadium next to a river" from an image may generate the two objects far apart given this caption. This disconnect indicates that spatial knowledge is encoded for perception but misaligned for generation.

We attribute this capability gap to how spatial information is introduced and passed through along the generation pathway. 
First, spatial relations are typically expressed implicitly and sparsely in language and are easily dominated by object-centric descriptions during generation.
In remote sensing imagery, dense object layouts make accurate modeling of spatial relations especially important for correct scene understanding and generation, thereby amplifying the impact of this limitation.
Second, in unified architectures that interface large language models with diffusion-based image generators, textual instructions are mapped into a compact conditioning representation prior to image synthesis. During this process, abstract and relational information—such as spatial layout—is more fragile than appearance-related cues and is therefore more susceptible to attenuation.

To address this gap, we propose Uni-RS, a unified multimodal model tailored for remote sensing imagery. Uni-RS incorporates three complementary design choices to improve spatial faithfulness during generation. First, Spatial-Layout Planning addresses the tendency of spatial constraints to be overshadowed by object-centric descriptions by reformulating textual instructions into structured layout plans that make spatial information more salient. Second, Spatial-Aware Query Supervision mitigates the loss of spatial information in the pathway from MLLM to diffusion model by explicitly supervising a subset of learnable queries to encode spatial relations in the instruction. Third, we introduce Image–Caption Spatial Layout Variation, a rotation-based data construction strategy that improves spatial adherence by training the model on the same objects arranged under different spatial layouts.

To summarize, our main contributions are fourfold:
\begin{itemize}
    \item We identify and systematically characterize the Spatial Reversal Curse in unified remote sensing models.
    \item We present Uni-RS, the first unified remote sensing multimodal model, which improves spatial faithfulness in image generation.
    \item The proposed Uni-RS achieves consistent improvements in spatially faithful image generation while maintaining strong multimodal understanding performance.
    \item We construct RS-Spatial, a large-scale spatial annotation dataset that enhances spatial faithfulness in unified model training. 
\end{itemize}

\section{Related Work}

\subsection{Unified Understanding and Generation}
Recent research has significantly advanced unified multimodal models capable of both image understanding and generation. Existing approaches can be broadly categorized into three paradigms. The first follows an autoregressive (AR) formulation, modeling both vision and language as a unified sequence within a single large language model (LLM)~\cite{team2024chameleon,wang2024emu3,chen2025janus}. The second paradigm also leverages a single MLLM for understanding and generation, but adopts a diffusion-based strategy for image synthesis~\cite{zhou2024transfusion,xie2025showo}. The third paradigm combines pretrained MLLMs with diffusion models by assigning semantic reasoning to the MLLM and delegating high-fidelity image generation to the diffusion component~\cite{pan2025transfer,wu2025qwen,chen2025blip3}.
Our approach falls into this third paradigm, extending it to a unified remote sensing setting with explicit spatial modeling.

\subsection{Unified Understanding and Generation in RS}

While unified understanding and generation have been extensively studied in general-domain vision--language models, research in the remote sensing (RS) community has largely followed decoupled trajectories. On the understanding side, recent years have witnessed the emergence of RS MLLM, beginning with RSGPT~\cite{hu2025rsgpt} and followed by instruction-tuned systems such as SkySenseGPT~\cite{luo2024skysensegpt}, GeoChat~\cite{kuckreja2024geochat}, GeoLLaVA-8K~\cite{wang2025geollavak}, and EarthGPT~\cite{zhang2024earthgpt}. In parallel, remote sensing image generation has progressed from early GAN-based approaches such as StrucGAN~\cite{zhao2021text} to diffusion-based generative models including DiffusionSat~\cite{khanna2024diffusionsat}, CRS-Diff~\protect\cite{tang2024crs}, Text2Earth~\protect\cite{liu2025text2earth}, and MetaEarth~\cite{yu2024metaearth}. Despite these advances, multimodal understanding and image generation in RS have largely been studied in isolation, and a unified framework that jointly supports understanding and faithful image generation remains underexplored.

\subsection{Spatial Planning in Vision--Language Models}

Chain-of-Thought (CoT) prompting~\cite{wei2022chain} and its multimodal extensions~\cite{zhang2024multimodal,liu2023visual} have been widely adopted to elicit step-by-step reasoning in vision--language models, enabling more interpretable decision-making and complex task decomposition. In RS domain, CoT-based supervised fine-tuning has further been explored to inject structured geographic reasoning into RS-VLMs, including approaches such as GeoChain~\cite{yerramilli2025GeoChain}, GeoSpatial CoT~\cite{liu2025towards}, GeoZero~\cite{wang2025geozero}, and GeoReason~\cite{li2026georeason}, as well as recent reasoning-centric models trained on large-scale RS reasoning datasets~\cite{TinyRS-R1,fiaz2025geovlm,li2025segearth}. 
However, these methods primarily focus on producing or supervising explicit reasoning traces for understanding tasks, rather than externalizing spatial layout information in a form that can directly guide image generation. In contrast, our work emphasizes spatial planning over reasoning explanation, aiming to make object layouts and spatial constraints explicit for faithful text-to-image synthesis.

\section{Characterizing the Spatial Reversal Curse}
\label{sec:empirical_study}
While unified MLLMs have demonstrated strong performance in general vision–language domains, their transfer to RS imagery exposes a non-trivial discrepancy. When adapting the unified model to RS data, we observe the Spatial Reversal Curse between spatial understanding and image generation: Although the model can accurately recognize spatial relations in the understanding direction, it often fails to faithfully execute the same relations in the generation direction.

Crucially, this discrepancy cannot be explained by object hallucination or low visual fidelity, as the target objects are often correctly synthesized but misplaced. To determine whether this phenomenon reflects isolated failure cases or a fundamental limitation of unified RS models, we move beyond qualitative examples and conduct a controlled quantitative characterization.

\subsection{A Symmetric Evaluation Protocol for Spatial Consistency}
To rigorously characterize the asymmetry between spatial understanding and spatial generation, we design a paired and symmetric evaluation protocol. We adopt RSIEval~\cite{hu2025rsgpt} as our testbed due to its suitability for spatial analysis, which provides high-resolution imagery with expert-annotated captions that explicitly describe spatial layouts and object positions.

To isolate the model’s ability to associate a single object with its explicit spatial location, we restrict our evaluation to unary spatial relations and adopt a standardized 3×3 grid as a shared spatial reference system for both understanding and generation tasks. The evaluation set is constructed through a model-assisted, human-verified pipeline, yielding 126 high-quality samples with unambiguous object–location annotations (see details in the supplementary materials).

We define complementary metrics to quantify this asymmetry. For spatial understanding, we formulate the task as a nine-way classification problem. Given an image \(I\) and the query “Where is the {object} located?”, the model is required to predict the correct grid cell. We report Spatial Understanding Accuracy (SUA), defined as:
\begin{equation}
\text{SUA} = \frac{1}{N} \sum_{i=1}^{N} \mathbf{1}(\hat{L}_i = L_{gt}^{(i)})
\label{eq:sua}
\end{equation}
where \(\hat{L}_i\) and \(L_{gt}^{(i)}\) denote the predicted and ground-truth locations for the \(i\)-th sample, respectively.

For image generation, we evaluate spatial faithfulness using two hierarchical metrics. The first is Object Generation Rate (OGR), which measures whether the model successfully synthesizes the target object, regardless of its spatial placement:
\begin{equation}
\text{OGR} = \frac{1}{N} \sum_{i=1}^{N} \mathbf{1}(E_i = 1),
\label{eq:ogr}
\end{equation}
where \(E_i = 1\) indicates that the target object is correctly detected in the generated image. OGR serves as an upper bound on spatial performance, as correct spatial placement is impossible if the object itself is missing.

The second metric is Spatially Faithful Rate (SFR), which measures the joint probability that the target object is both generated and placed in the correct spatial region specified by the prompt:
\begin{equation}
\text{SFR} = \frac{1}{N} \sum_{i=1}^{N} \mathbf{1}(E_i = 1 \land P_i = 1),
\label{eq:sfr}
\end{equation}
where \(P_i = 1\) denotes correct placement in the ground-truth grid cell.
\begin{figure}
    \centering
    \includegraphics[width=0.8\linewidth]{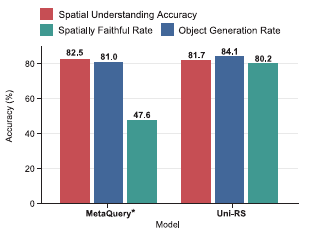}
    \caption{Quantitative characterization of the Spatial Reversal Curse. The baseline MetaQuery* model exhibits a pronounced gap between spatial understanding and spatially faithful text-to-image generation. The * denotes that the model was trained by us under the same experimental settings as described in Section 5.1 due to the absence of public checkpoints. The Uni-RS largely reduces this gap after incorporating the proposed spatial mechanisms.}
    \label{fig:reversal_results}
\end{figure}

\begin{figure*}[t]
    \centering
    \includegraphics[width=\linewidth]{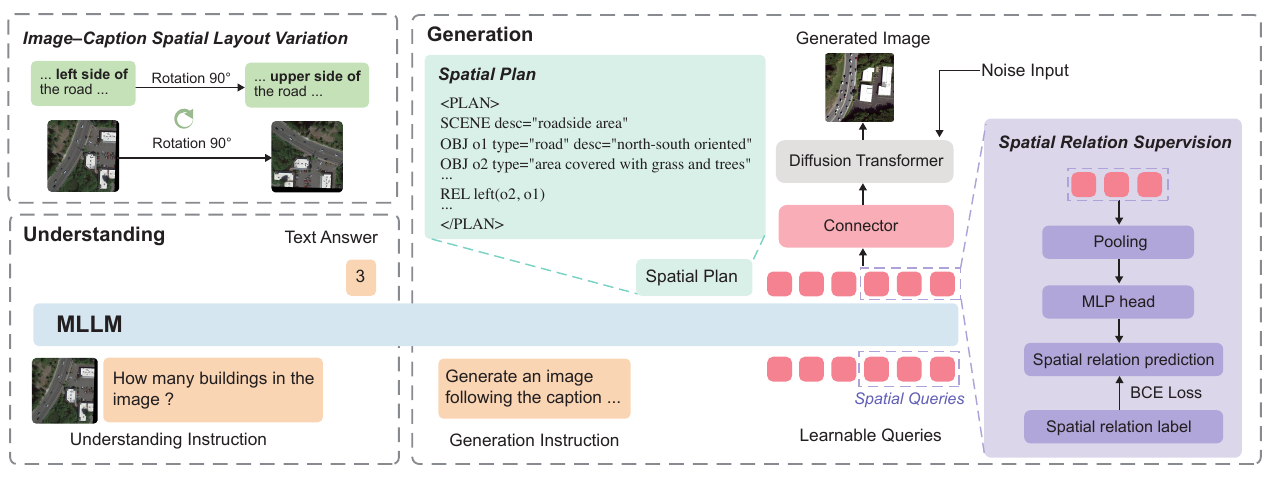}
    \caption{Overview of the Uni-RS framework. Uni-RS integrates a multimodal large language model (MLLM) with a diffusion-based image generator through learnable queries and a Transformer-based connector. During training, the framework is further enhanced by Spatial-Layout Planning, Spatial-Aware Query Supervision, and Image–Caption Spatial Layout Variation, which jointly reinforce instruction-level spatial constraints and improve spatial faithfulness during generation.}
    \label{fig:framework}
\end{figure*}

\subsection{Findings}
We tested the Spatial Reversal Curse using MetaQuery. As no public checkpoint is available, we train MetaQuery using the official implementation under the same experimental settings described in Section 5.1. As shown in Figure \ref{fig:reversal_results}, MetaQuery achieves a high Spatial Understanding Accuracy (SUA) of 82.54\%, which indicates that the model possesses sufficient perceptual capacity to recognize spatial layouts from remote sensing imagery in the understanding direction.

In contrast, when conditioned on equivalent textual instructions for text-to-image generation, the same model exhibits a pronounced degradation in spatial faithfulness. Although the target object is frequently synthesized, with an Object Generation Rate (OGR) of 80.95\%, correct spatial placement is achieved in less than half of the cases, as reflected by a Spatially Faithful Rate (SFR) of 47.62\%.

The discrepancy yields a substantial spatial gap of \[
\Delta = \text{OGR} - \text{SFR} = 33.33\%,
\]
indicating that generation failures predominantly stem from incorrect spatial execution rather than object hallucination or missing content.

To further examine the generality of this phenomenon beyond the MetaQuery-style paradigm, we additionally experiment with Show-o~\cite{xie2025showo}. After fine-tuning Show-o on remote sensing data, we observe a similar asymmetry, where $\Delta$ reaches 27.78\%. Detailed experimental settings and results for Show-o are provided in the supplementary materials.

\section{Methodology}
\label{sec:methodology}

\subsection{Model Architecture}
\label{sec:methodology_overall}

We adopt a unified framework that bridges a multimodal large language model (MLLM) and a diffusion-based image generator through a set of learnable queries. As illustrated in Fig.~\ref{fig:framework}, textual instructions are first processed by the MLLM, and the resulting query representations are transformed into conditioning signals that guide diffusion-based image synthesis. We adopt Qwen2.5-VL-3B~\cite{Qwen2.5-VL} as the multimodal large language model backbone and SANA-1.6B~\cite{xie2024sana} as the diffusion-based image generator, bridged by \(N=256\) learnable queries and a 24-layer Transformer-based connector. This configuration follows the best practice of MetaQuery~\cite{pan2025transfer} and is fixed across all experiments. 

Within this unified framework, Uni-RS incorporates three complementary spatial mechanisms. \emph{Spatial-Layout Planning} rewrites the input instruction into a structured spatial plan, which is attended by learnable queries to pass the spatial layout through to diffusion-based synthesis. \emph{Spatial-Aware Query Supervision} further supervises a designated subset of learnable queries with spatial relation labels through an auxiliary prediction head. \emph{Image--Caption Spatial Layout Variation} augments training with rotated image--caption pairs to strengthen the model's sensitivity to spatial layouts beyond object identity.

\subsection{Spatial-Layout Planning}
Remote sensing image generation often involves complex scenes with dense object distributions.
In such cases, spatial layout constraints are easily overshadowed by long, object-centric descriptions:
object names correspond to salient visual patterns and receive strong supervision,
whereas spatial relations are abstract and only implicitly learned through standard generative objectives.

To address this, we introduce Spatial-Layout Planning, which rewrites a user instruction
into a structured layout specification expressed in text.
Specifically, instead of conditioning on raw captions alone, the MLLM first generates
a plan enclosed by a \texttt{<PLAN>} tag that decomposes the instruction into three
structured components: the global scene context (SCENE), object instances with coarse
spatial locations (OBJ), and explicitly stated spatial relations (REL),
as illustrated in Figure~\ref{fig:framework}.
This structured plan makes spatial constraints---which are otherwise buried in
free-form descriptions---directly visible to the learnable queries, encouraging them
to better preserve the intended layout during image generation.

We note that, although Spatial-Layout Planning produces an auxiliary textual plan like chain-of-thought (CoT), its purpose differs fundamentally from CoT prompting: rather than explaining reasoning steps, the plan explicitly externalizes spatial layout information that is otherwise implicit in natural language instructions.

\subsection{Spatial-Aware Query Supervision}

The learnable queries serve as the primary carriers of conditioning information passed from the MLLM to the diffusion model. When rich language instructions are compressed into a fixed set of queries, fine-grained spatial cues may be attenuated. To mitigate this effect, we introduce a spatial-aware query supervision term: during training, a subset of queries, denoted as \(\mathbf{Q}_{\text{spa}} \subset \mathbf{Q}\), is softly biased to retain spatial relations that are explicitly mentioned in the instruction via an auxiliary supervision signal.

This supervision relies on a closed-set vocabulary of spatial relations that reflects how spatial concepts are expressed in remote sensing captions. Drawing on cognitive geography, we organize spatial expressions into five broad categories: directional, topological, distance-based, orientation-based, and distributional relations, with the last capturing characteristic multi-object arrangements in overhead imagery. Based on this taxonomy, we construct a spatial relation vocabulary \(\mathcal{V}\) with \(K=79\) spatial relation labels, which provides coverage of spatial expressions appearing in the training corpus (see supplementary materials for the full spatial relation vocabulary).

Given an instruction, we extract the spatial relations it explicitly mentions to obtain a multi-hot target vector \(\mathbf{y} \in \{0,1\}^{K}\). During training, features from \(\mathbf{Q}_{\text{spa}}\) are aggregated (e.g., via mean pooling) and passed through a lightweight projection head to predict \(\hat{\mathbf{y}} \in \mathbb{R}^{K}\):
\begin{equation}
\hat{\mathbf{y}} = \sigma\!\left(\mathrm{MLP}\!\left(\mathrm{Pool}(\mathbf{Q}_{\text{spa}})\right)\right).
\end{equation}
We supervise this prediction using a binary cross-entropy loss:
\begin{equation}
\mathcal{L}_{\text{spa}} = \mathrm{BCE}(\hat{\mathbf{y}}, \mathbf{y}).
\end{equation}
This auxiliary loss biases the spatial-aware queries to preserve instruction-level spatial cues that are subsequently reflected in the conditioning signals used for image generation. In our experiments, we designate the last $32$ queries as $\mathbf{Q}_{\text{spa}}$, providing sufficient capacity to encode spatial relations, while preserving the remaining queries for modeling object semantics and other non-spatial information.

\subsection{Image–Caption Spatial Layout Variation}


Finally, we introduce Image–Caption Spatial Layout Variation as a geometry-aware training strategy. This design exploits a geometric property of overhead imagery: rotating a remote sensing image preserves scene semantics, object categories and attributes, as well as topological, distance-based, and distributional spatial relations, while resulting in systematic changes to absolute object locations and orientation- or direction-dependent spatial relations.

Concretely, for each training image, we generate rotated variants with angles \(\theta \in \{90^\circ, 180^\circ, 270^\circ\}\). To maintain geometric consistency between vision and language, the associated caption is rewritten accordingly. Given a rotation angle \(\theta\), a large language model is prompted to identify spatial expressions in the original caption and update those that are direction- or orientation-dependent based on the corresponding geometric transformation (e.g., changing ``North of" to ``East of" for a $90^\circ$ clockwise rotation), while preserving scene descriptions, object categories, attributes, and rotation-invariant spatial relations. The rewritten caption remains linguistically fluent and semantically aligned with the rotated image.

As a result, the training data contain paired image–caption instances that depict identical scenes and objects but differ in absolute object locations and orientation-sensitive spatial relations. This paired variation isolates spatial layout changes, enabling the model to associate spatial-related descriptions with geometric transformations instead of memorizing absolute object locations.

\begin{table}[t]
\centering
\small
\setlength{\tabcolsep}{8pt}
\begin{tabular}{lccc}
\toprule
Method 
& FID ($\downarrow$) 
& \makecell{Zero-Shot\\Cls-OA ($\uparrow$)} 
& CLIP ($\uparrow$) \\
\midrule
Attn-GAN                            & 95.81  & 32.56\%  & 20.19 \\
DAE-GAN                               & 93.15  & 29.74\%  & 19.69 \\
DF-GAN                                & 109.41 & 51.99\%  & 19.76 \\
Lafite                                & 74.11  & 49.37\%  & 22.52 \\
DALL-E                                & 191.93 & 28.59\%  & 20.13 \\
Txt2Img-MHN$_{\text{VQVAE}}$          & 175.36 & 41.46\%  & 21.35 \\
Txt2Img-MHN$_{\text{VQGAN}}$          & 102.44 & 65.72\%  & 20.27 \\
RSDiff                                & 66.49  & --       & --    \\
CRS-Diff                              & 50.72  & 69.31\%  & 20.33 \\
Text2Earth$^\dagger$                  & 67.92  & 57.74\%       & 23.91 \\
Text2Earth                            & 24.49  & 90.26\%  & 25.62 \\
\midrule
Uni-RS (Ours)$^\dagger$               & 46.83  & 66.93\%  & 24.11 \\
Uni-RS (Ours)                         & \textbf{22.11} & \textbf{91.63\%} & \textbf{25.68} \\
\bottomrule
\end{tabular}
\caption{Comparison with existing text-to-image generation methods on RSICD.
$\dagger$ indicates methods evaluated without fine-tuning on RSICD. See supplementary materials for detailed descriptions of baselines.}
\label{tab:rsicd}
\end{table}

\begin{table}[!h]
\centering
\small
\setlength{\tabcolsep}{3.5pt}
\begin{tabular}{lccc|cc}
\toprule
& \multicolumn{3}{c}{RSIEval} & \multicolumn{2}{c}{VRSBench} \\
\cmidrule(lr){2-4} \cmidrule(lr){5-6}
Method 
& FID ($\downarrow$) 
& SFR ($\uparrow$) 
& CLIP ($\uparrow$) 
& FID ($\downarrow$) 
& CLIP ($\uparrow$) \\
\midrule
Text2Earth$^\dagger$                      
& 17.07 & 63.40 & 23.74 
& 8.28  & 24.85 \\

Text2Earth        
& 16.22 & 65.10 & 23.79 
& 7.42  & 24.93 \\
\midrule
Uni-RS (Ours)                 
& \textbf{13.04} & \textbf{80.16} & \textbf{24.01} 
& \textbf{4.51}  & \textbf{25.46} \\
\bottomrule
\end{tabular}
\caption{Comparison of spatial fidelity and generation quality on RSIEval and VRSBench.
$\dagger$ indicates methods evaluated without fine-tuning on the corresponding dataset.}
\label{tab:rsieval_vrsbench_combined}
\end{table}

\begin{table*}[!h]
\centering
\small
\setlength{\tabcolsep}{3.9pt}
\renewcommand{\arraystretch}{1.05}
\begin{tabular}{lccccc|lccccccc}
\toprule
\multicolumn{6}{c|}{\textbf{RSIEval}} 
& \multicolumn{8}{c}{\textbf{VRSBench-val}} \\

\cmidrule(lr){1-6} \cmidrule(lr){7-14}

& \multicolumn{4}{c}{\textbf{Captioning}} & \textbf{VQA} & 
& \multicolumn{4}{c}{\textbf{Captioning}} & \textbf{VQA} & \multicolumn{2}{c}{\textbf{Grounding}} \\

\cmidrule(lr){2-5} \cmidrule(lr){6-6}
\cmidrule(lr){8-11} \cmidrule(lr){12-12} \cmidrule(lr){13-14}

\textbf{Method} 
& B-4 & METEOR & R-L & CIDEr & Avg
& \textbf{Method}
& B-4 & METEOR & R-L & CIDEr & Avg
& Acc@0.5 & Acc@0.7 \\

\midrule
BLIP2 
& 0.0 & 3.4 & 11.6 & 0.1 & 45.6
& Mini-Gemini
& 14.3 & 21.5 & 36.8 & 33.5 & 77.8
& 30.1 & 6.8 \\

MiniGPT4 
& 17.3 & 19.9 & 35.3 & 16.3 & 21.8
& GeoChat
& 13.8 & 21.1 & 35.2 & 28.2 & 76.0
& 49.8 & 19.9 \\

RSGPT 
& 22.1 & 23.6 & 41.7 & 31.4 & 65.2
& GeoPix
& 14.0 & 23.4 & 36.3 & 31.3 & 74.9
& 49.8 & 20.0 \\

GeoPix$^{\dagger}$ 
& 2.8 & 9.4 & 16.1 & 2.9 & 65.5
& LLaVA-1.5
& 14.7 & 21.9 & 36.9 & 33.9 & 76.4
& 41.6 & 13.6 \\

\midrule
\textbf{Ours} 
& 16.8 & 20.9 & 38.8 & 22.3 & 65.4
& \textbf{Ours}
& 12.2 & 20.7 & 34.6 & 30.0 & 74.0
& 45.7 & 13.3 \\

\bottomrule
\end{tabular}
\caption{Unified comparison on RSIEval and VRSBench-val across multimodal understanding tasks.
$^{\dagger}$ indicates methods evaluated without task-specific fine-tuning.}
\label{tab:unified_understanding}
\end{table*}

\subsection{RS-Spatial: A Spatial Annotation Dataset for Training Uni-RS}

To support the proposed spatial mechanisms in a unified and systematic manner, 
we construct a new large-scale spatial annotation dataset, termed RS-Spatial, built upon RSICap~\cite{hu2025rsgpt} and VRSBench~\cite{li2024vrsbench}. This dataset aims to provide additional spatial layout supervision for remote sensing text-to-image generation in unified models.

RS-Spatial contains three types of spatial annotations:
(i) structured layout descriptions for Spatial-Layout Planning,
(ii) spatial relation labels aligned with the predefined vocabulary for spatial-aware query supervision, and
(iii) rewritten captions corresponding to rotated images for image–caption spatial layout variation.
All annotations are generated using Qwen3-VL-235B-A22B-Instruct~\cite{Qwen3-VL} through a shared two-stage pipeline. In the first stage, the model is prompted to produce task-specific annotations for each image–caption pair. In the second stage, the same model verifies the consistency between the generated annotations and the original image–caption pairs or applied rotations, and samples that fail this verification are discarded.
In total, RS-Spatial contains 20,496 structured spatial layout plans, 22,849 spatial relation annotations, and 36,702 rotation-consistent image–caption pairs.

Importantly, RS-Spatial introduces no new semantic content beyond what is explicitly supported by the original captions, ensuring that all spatial relations and layout elements are strictly text-faithful. As a result, the dataset provides a unified spatial supervision resource that supports spatial planning, spatial relation modeling, and rotation-based data augmentation within a single framework. The dataset will be publicly released.

\section{Experiments}
\subsection{Experimental Setup}
\paragraph{Training Protocol.}
We adopt a two-stage training protocol, in which Stage~1 optimizes only the text-to-image generation objective, whereas Stage~2 jointly optimizes both understanding and generation tasks.

\paragraph{Stage~1 (Generation Training).}
In the first stage, we freeze the MLLM and update only the Transformer connector and the diffusion backbone. The model is trained on Git10M~\cite{liu2025text2earth}, a large-scale remote sensing image--text corpus containing approximately 10 million pairs. This stage focuses on text-to-image generation, enabling the diffusion backbone to internalize the visual characteristics of overhead imagery, including top-down viewpoints and remote-sensing-specific textures, thereby establishing a robust generative prior.

\paragraph{Stage~2 (Joint Understanding--Generation Training).}
In the second stage, we unfreeze the remaining trainable components while keeping the MLLM visual encoder (ViT) and the diffusion VAE fixed. We jointly optimize multimodal understanding objectives---including VQA, captioning, and grounding---together with text-to-image generation. Training data in this stage include RSICap~\cite{hu2025rsgpt}, VRSBench~\cite{li2024vrsbench}, and the proposed RS-Spatial dataset. RSICap and VRSBench are used for multimodal understanding, with their captions reused as prompts for text-to-image generation, while RS-Spatial provides spatial layout plans, spatial relation labels, and rotation-consistent image--caption pairs to support spatial planning, spatial relation modeling, and geometry-consistent data augmentation. Note that for spatial layout plans, we apply them to both understanding and generation training, enabling the model to learn not only to produce layout plans from language inputs, but also to utilize these plans for image synthesis.

\paragraph{Implementation Details.}
All models are trained using the AdamW optimizer with a batch size of 1024 on 8 NVIDIA A800 GPUs (80GB). Both training stages employ a cosine learning rate schedule with a 10\% warmup phase. Stage~1 is trained for 10 epochs with an initial learning rate of $2\times10^{-5}$. Stage~2 is fine-tuned for 5 epochs with an initial learning rate of $1\times10^{-5}$. The balancing coefficient for the spatial supervision loss is set to $\lambda = 0.1$.
\subsection{Comparison with SoTA Models}

\label{sec:sota}
We compare Uni-RS with existing remote sensing text-to-image generation models on RSICD, RSIEval, and VRSBench.

\paragraph{Text-to-Image Generation on RSICD.}
Table~\ref{tab:rsicd} reports results on RSICD using FID, zero-shot classification accuracy (Cls-OA), and CLIP score.
Following common practice in prior work, we report results under two settings: with RSICD fine-tuning, where Uni-RS is fine-tuned on the RSICD training set for five epochs before evaluation, and without RSICD fine-tuning.

As shown in Table~\ref{tab:rsicd}, fine-tuning Uni-RS on RSICD leads to state-of-the-art performance across all three metrics.
Notably, even without RSICD fine-tuning, Uni-RS outperforms Text2Earth and several prior methods that are explicitly fine-tuned on RSICD, such as CRS-Diff~\protect\cite{tang2024crs} and Txt2Img-MHN~\cite{xu2023txt2img}.
These results demonstrate strong text-to-image generation capability and generalization of Uni-RS, consistent with prior findings that enhanced multimodal understanding can benefit generation performance.

\begin{figure*}[h]
    \centering
    \includegraphics[width=\linewidth]{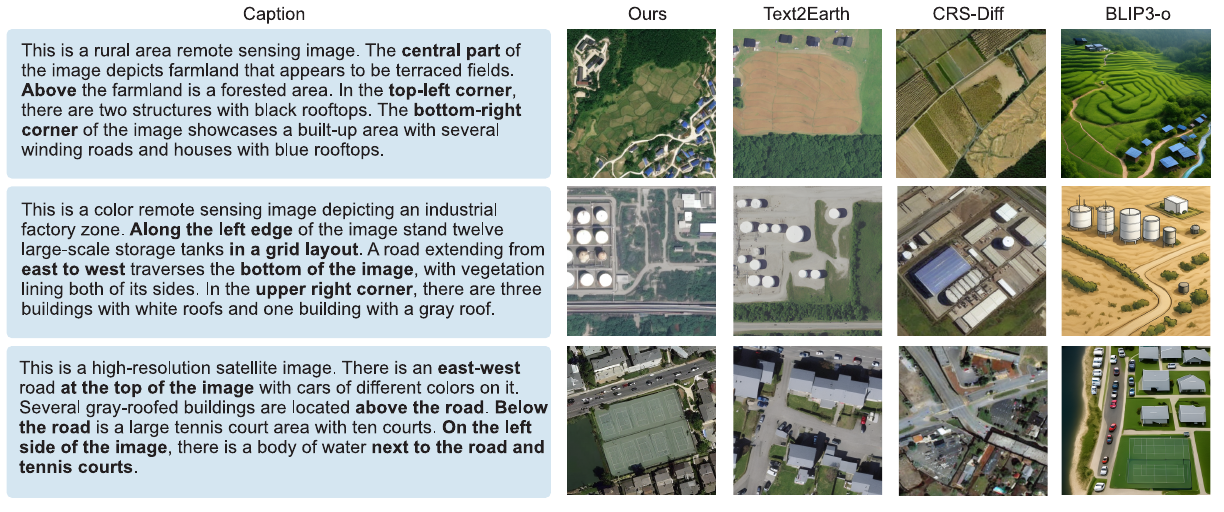}
    \caption{Qualitative comparison of spatial faithfulness in text-to-image generation. BLIP3-o~\protect\cite{chen2025blip3} is a general-domain unified multimodal model for understanding and generation, while CRS-Diff~\protect\cite{tang2024crs} and Text2Earth~\protect\cite{liu2025text2earth} are remote-sensing-specific text-to-image generation models. The Text2Earth model is fine-tuned using RSICap and VRSBench datasets to produce $512\times512$ images.}
    \label{fig:qualitative}
\end{figure*}
\paragraph{Spatial Faithful Generation on RSIEval and VRSBench.}

Table~\ref{tab:rsieval_vrsbench_combined} reports results on RSIEval and VRSBench. Compared to RSICD, captions in RSIEval and VRSBench are longer and more structured, typically involving multiple objects and explicitly specified spatial relations.
These characteristics make spatial layout adherence a central factor in evaluating generation quality. For a fair comparison with the prior state-of-the-art model Text2Earth, we further fine-tune Text2Earth on VRSBench and RSICap for 5 epochs using the same training protocol.

As shown in Table~\ref{tab:rsieval_vrsbench_combined}, Uni-RS outperforms the fine-tuned Text2Earth model across all metrics on both benchmarks. On RSIEval, Uni-RS reduces FID by approximately 20\% compared to the fine-tuned Text2Earth model, while improving the Spatially Faithful Rate (SFR) by approximately 23\%. On VRSBench, Uni-RS achieves a 39\% reduction in FID and an approximately 2\% increase in CLIP score. The pronounced improvement in SFR indicates that Uni-RS is not only capable of generating the target objects, but also places them in accordance with the layout specified in the caption. 

\paragraph{Understanding Tasks.}
Table~\ref{tab:unified_understanding} reports results on image captioning, visual question answering, and visual grounding tasks.
Across all three tasks, Uni-RS achieves performance comparable to existing remote sensing vision--language models, with minimal performance gaps on the reported metrics. These results demonstrate that Uni-RS functions as an effective unified model, exhibiting strong multimodal understanding while simultaneously supporting text-to-image generation.

\paragraph{Qualitative Analysis.}
\label{sec:qualitative}

As shown in Figure \ref{fig:qualitative}, qualitative results further corroborate the quantitative findings. General-domain unified models are able to correctly identify intended objects, while they often fail to render the overhead perspective. In contrast, remote-sensing-specific generation models produce images with more realistic overhead appearances, but frequently struggle to faithfully follow the objects and to place them in spatial configurations consistent with the described layouts. By comparison, Uni-RS is able to consistently synthesize the objects described in the caption while simultaneously respecting their intended spatial arrangements.

\subsection{Ablation Study}

\label{sec:ablation}
We perform an ablation study to evaluate the contribution of each proposed component.
Results are reported in Table~\ref{tab:ablation} on RSIEval and VRSBench.
Overall, the ablation results show that all three components consistently improve spatial faithfulness, and their effects accumulate in a stable and complementary manner. Introducing spatial layout planning already leads to a clear gain in spatially faithful generation, while simultaneously improving image fidelity and semantic alignment, as reflected by reduced FID and higher CLIP scores. Image–caption spatial layout variation further strengthens this trend by exposing the model to geometry-consistent variations, yielding additional improvements in both spatial adherence and overall generation quality. In contrast, spatial-aware query supervision primarily enhances spatial faithfulness by reinforcing instruction-level spatial constraints in the conditioning space, while having only a marginal impact on FID and CLIP. When combined, the three components progressively mitigate the Spatial Reversal Curse, achieving the highest spatial fidelity across benchmarks without degrading generation quality.

\begin{table}[t]
\centering
\small
\setlength{\tabcolsep}{4.5pt}
\begin{tabular}{lcc|ccc}
\toprule
& \multicolumn{2}{c}{VRSBench} & \multicolumn{3}{c}{RSIEval} \\
\cmidrule(lr){2-3} \cmidrule(lr){4-6}
Method 
& FID 
& CLIP 
& FID 
& CLIP 
& \textbf{SFR} \\
\midrule
Base 
& 4.91  & 25.33
& 13.47 & 23.70 & 47.62 \\

Base + SLP
& 4.62  & 25.40 
& 13.44 & 23.81 & 54.00 \\

Base + SLP + SQS
& 4.64  & 25.42 
& 13.33 & 23.79 & 63.49 \\

Base + SLP + SQS + ISV
& \textbf{4.51}  & \textbf{25.46} 
& \textbf{13.04} & \textbf{24.01} & \textbf{80.16} \\
\bottomrule
\end{tabular}
\caption{Ablation study on VRSBench and RSIEval.
Components are added cumulatively:
(1) Spatial-Layout Planning (SLP),
(2) Spatial-Aware Query Supervision (SQS),
(3) Image--Caption Spatial Layout Variation (ISV).}
\label{tab:ablation}
\end{table}

\section{Conclusion}
We presented Uni-RS, the first unified remote sensing vision--language model that systematically investigates and mitigates the Spatial Reversal Curse between understanding and generation.
By introducing spatially explicit mechanisms through Spatial-Layout Planning and incorporating spatial supervision for latent queries, Uni-RS improves spatial faithfulness in text-to-image generation while preserving strong multimodal understanding. Extensive experiments across multiple benchmarks demonstrate that Uni-RS effectively integrates spatial understanding and spatial execution within a single unified framework.

\clearpage

\section*{Acknowledgements}
We acknowledge the support of the National Natural Science Foundation of China (Grant Nos. 42430106 and 42401407) and the China Postdoctoral Science Foundation (Grant No. 2024M760086).

\section*{Contribution Statement}
Weiyu Zhang and Yuan Hu contributed equally to this work. Yu Liu is the corresponding author.

\bibliographystyle{named}
\bibliography{ijcai26}

\end{document}